\documentclass[11pt,a4paper]{article}
\usepackage[hyperref]{acl2019}
\usepackage{times}
\usepackage{latexsym}

\usepackage{url}
\usepackage{algorithm}
\usepackage[noend]{algpseudocode}
\usepackage{amsmath,amsfonts,amsthm,amssymb} 

\usepackage{booktabs}
\usepackage{blindtext}
\usepackage{enumitem}
\usepackage{tabularx}
\usepackage{pgfplots}
\pgfplotsset{compat=newest}
\usepackage{pgf-pie}
\usepackage{mwe}
\usepackage{subfig}
\usepackage{soul}
\usepackage{graphicx}

\aclfinalcopy 

\algrenewcommand\algorithmicindent{0.6em}%
\algnewcommand{\LineComment}[1]{\State {\small // #1}}

\title{Revisiting Joint Modeling of Cross-document\\Entity and Event Coreference Resolution}

\author{Shany Barhom$^1$, Vered Shwartz$^1$, Alon Eirew$^2$, Michael Bugert$^3$, Nils Reimers$^3$, \and Ido Dagan$^1$ \\
\mbox{}\\
$^1$ Computer Science Department, Bar-Ilan University\\
$^2$ Intel AI Lab, Israel \\
$^3$ Ubiquitous Knowledge Processing Lab, Technische Universitat Darmstadt, Germany \\
\small{\texttt{\{shanyb21,vered1986\}@gmail.com},~~~\texttt{alon.eirew@intel.com}}\\
\small{~\texttt{\{bugert,reimers\}@ukp.informatik.tu-darmstadt.de},~~~\texttt{dagan@cs.biu.ac.il}}}

\date{}

\begin{document}
\maketitle
\begin{abstract}
  Recognizing coreferring events and entities \textit{across} multiple texts is crucial for many NLP applications. Despite the task's importance, research focus was given mostly to within-document entity coreference, with rather little attention to the other variants. We propose a neural architecture for cross-document coreference resolution. Inspired by \newcite{lee-EtAl:2012:EMNLP-CoNLL}, we jointly model entity and event coreference. We represent an event (entity) mention using its lexical span, surrounding context, and relation to entity (event) mentions via predicate-arguments structures. Our model outperforms the previous state-of-the-art event coreference model on ECB+, while providing the first entity coreference results on this corpus. Our analysis confirms that all our representation elements, including the mention span itself, its context, and the relation to other mentions contribute to the model's success.

\end{abstract}

\section{Introduction}
\label{sec:intro}
Recognizing that various textual spans
across multiple texts refer to the same entity or event is
an important NLP task. For example, consider the following news headlines: 

\vspace{-7pt}
\begin{table}[!h]
    \small
    \begin{tabular}{|l|}
        \hline
         1. \textit{\textbf{2018 Nobel prize for physics} \underline{goes to} \textbf{Donna Strickland}} \\
         2.  \textit{\textbf{Prof. Strickland} \underline{is awarded} \textbf{the Nobel prize for physics}} \\
         \hline
    \end{tabular}
\end{table}
\vspace{-7pt}

\noindent Both sentences refer to the same entities (Donna Strickland and the Nobel prize for physics) and the same event (awarding the prize), using different words. In coreference resolution, the goal is to cluster expressions that refer to the same entity or event in a text, whether within a single document or across a document collection. Recently, there has been increasing interest in cross-text inferences, for example in question answering \cite{welbl2018constructing,yang-EtAl:2018:EMNLP1,khashabi-EtAl:2018:N18-1,postma-ilievski-vossen:2018:S18-1}. Such applications would benefit from effective cross-document coreference resolution. 

Despite the importance of the task, the focus of most coreference resolution research has been on its within-document variant, and rather little on cross-document coreference (CDCR). The latter is sometimes addressed partially using entity linking, which links mentions of an entity to its knowledge base entry. However, cross-document entity coreference is substantially broader than entity linking, addressing also mentions of common nouns and unfamiliar named entities. 

The commonly used dataset for CDCR is ECB+ \cite{cybulska2014using}, which annotates within-document coreference as well. The annotations are denoted separately for entities and events, making it possible to solve one task while ignoring the other. Indeed, to the best of our knowledge, all previously published work on ECB+ addressed only event coreference. 

Cross-document entity coreference has been addressed on EECB, a predecessor of the ECB+ dataset. \newcite{lee-EtAl:2012:EMNLP-CoNLL} proposed to model the entity and event coreference tasks jointly, leading to improved performance on both tasks. Their model preferred to cluster event mentions whose arguments are in the same entity coreference cluster, and vice versa. For instance, in the example sentences above, a system focusing solely on event coreference may find it difficult to recognize that \textit{goes to} and \textit{awarded} are coreferring, while a joint model would leverage the coreference between their arguments. 

Inspired by the success of the joint approach of \newcite{lee-EtAl:2012:EMNLP-CoNLL}, we propose a joint neural architecture for CDCR. In our joint model, an event (entity) mention representation is aware of other entities (events) that are related to it by predicate-argument structure. We cluster mentions based on a learned pairwise mention coreference scorer. 

A disjoint variant of our model, on its own, improves upon the previous state-of-the-art for event coreference on the ECB+ corpus \cite{kenyondean-cheung-precup:2018:S18-2} by 9.5 CoNLL $F_1$ points. To the best of our knowledge, we are the first to report performance on the entity coreference task in ECB+. 

Our joint model further improves performance upon the disjoint model by 1.2 points for entities and 1 point for events (statistically significant with $p < 0.001$). Our analysis further shows that each of the mention representation components contributes to the model's performance.\footnote{The code is available at \url{https://github.com/shanybar/event_entity_coref_ecb_plus}.}

\section{Background and Related Work}
\label{sec:background}
Coreference resolution is the task of clustering text spans that refer to the same entity or event. Variants of the task differ on two axes: (1) resolving entities (``\textit{Duchess of Sussex}'', ``\textit{Meghan Markle}'', ``\textit{she}'') vs. events (``\textit{Nobel prize for physics [goes to] Donna Strickland}'', ``\textit{Donna Strickland [is awarded] the 2018 Nobel prize for physics}''), and (2) whether coreferring mentions occur within a single document (WD: \emph{within-document}) or across a document collection (CD: \emph{cross-document}). 

\subsection{Datasets}
\label{sec:bg_datasets}

The largest datasets that include WD and CD coreference annotations for both entities and events are EECB \cite{lee-EtAl:2012:EMNLP-CoNLL} and ECB+ \cite{cybulska2014using}. Both are extensions of the Event Coreference Bank (ECB) \cite{bejan-harabagiu:2010:ACL} which consists of documents from Google News clustered into topics and annotated for event coreference. Entity coreference annotations were first added in EECB, covering both common nouns and named entities. 

ECB+ increased the difficulty level by adding a second set of documents for each topic (sub-topic), discussing a different event of the same type (\textit{Tara Reid enters a rehab center} vs. \textit{Lindsay Lohan enters a rehab center}). The annotation is not exhaustive, where only a number of salient events and entities in each topic are annotated.

\subsection{Models}
\label{sec:bg_models}

\paragraph{Entity Coreference.} Of all the coreference resolution variants, the most well-studied is WD entity coreference resolution  \cite[e.g.][]{durrett-klein:2013:EMNLP,clark-manning:2016:P16-1}. The current best performing model is a neural end-to-end system which considers all spans as potential entity mentions, and learns distributions over possible antecedents for each \cite{lee-EtAl:2017:EMNLP2017}. CD entity coreference has received less attention \cite[e.g.][]{Bagga:1998:ECC:980845.980859,rao-mcnamee-dredze:2010:POSTERS,TACL522}, often addressing the narrower task of entity linking, which links mentions of known named entities to their corresponding knowledge base entries \cite{shen2015entity}. 

\paragraph{Event Coreference.} Event coreference is considered a more difficult task, mostly due to the more complex structure of event mentions. While entity mentions are mostly noun phrases, event mentions may consist of a verbal predicate (\textit{acquire}) or a nominalization (\textit{acquisition}), where these are attached to arguments, including event participants and spatio-temporal information. 

Early models employed lexical features (e.g. head lemma, WordNet synsets, word embedding similarity) as well as structural features (e.g. aligned arguments) to compute distances between event mentions and decide whether they belong to the same coreference cluster \cite[e.g.][]{bejan-harabagiu:2010:ACL,bejan2014unsupervised,TACL634}. 

More recent work is based on neural networks.
\newcite{choubey-huang:2017:EMNLP20172} alternate between WD and CD clustering, each step relying on previous decisions. The decision to link two event mentions is made by the pairwise WD and CD scorers. Mention representations rely on pre-trained word embeddings, contextual information, and features related to the event's arguments. 

\newcite{kenyondean-cheung-precup:2018:S18-2} similarly encode event mentions using lexical and contextual features. Differently from \newcite{choubey-huang:2017:EMNLP20172}, they do not cluster documents to topics as a pre-processing step. Instead, they encode the document as part of the mention representation.

Most of the recent models were trained and evaluated on the ECB+ corpus, addressing solely the event coreference aspect of the dataset.

\paragraph{Joint Modeling.} Some of the prior models leverage the event arguments to improve their coreference decisions \cite{TACL634,choubey-huang:2017:EMNLP20172}, but mostly relying only on lexical similarity between arguments of candidate event mentions. A different approach was proposed by \newcite{lee-EtAl:2012:EMNLP-CoNLL}, who jointly predicted event and entity coreference. 

At the core of their model lies the assumption that arguments (i.e. entity mentions) play a key role in describing an event, therefore, knowing that two arguments are coreferring is useful for finding coreference relations between events, and vice versa. They incrementally merge entity or event clusters, computing the merge score between two clusters by learning a linear regression model based on discrete features. 

\newcite{lee-EtAl:2012:EMNLP-CoNLL} evaluated their model on EECB, outperforming disjoint CD coreference models for both entities and events. Nonetheless, as opposed to the more recent models, their representations are sparse. Lexical features are based on lexical resources such as WordNet \cite{miller1995wordnet}, which are limited in coverage, and context is modeled using semantic role dependencies, which often do not cover the entire sentential context. We revisit the joint modeling approach, trying to overcome prior limitations by using modern neural techniques, which provide better and more generalizable representations.

\section{Model}
\label{sec:model}
We propose an iterative algorithm that alternates between interdependent entity and event clustering, incrementally constructing the final clustering configuration. A single iteration for events is as follows (entity clustering is symmetric). We start by computing the mention representations (Section~\ref{sec:representation}), which couple the entity and event clustering processes. When predicting event clusters, the event mention representations are updated to consider the current configuration of entity clusters. The mention representations are then fed to an event mention pair scorer that predicts whether the mentions belong to the same cluster (Section~\ref{sec:scoring}). Finally, we apply agglomerative clustering where the cluster merging score is based on the predicted pairwise mention scores. Sections~\ref{sec:inference} and \ref{sec:training} detail the specifics of the inference and training procedures, respectively. Various implementation details are mentioned in Section~\ref{sec:implementation_details}. 

\subsection{Mention Representation}
\label{sec:representation}

Given a mention $m$ (entity or event), we compute a vector representation with the following features.

\paragraph{Span.} We combine word-level and character-level features. We compute word-level representations using pre-trained word embeddings. For events, we take the embedding of the head word, while for entities we average over the mention's words. Character-level representations are complementary, and may help with out-of-vocabulary words and spelling variations. We compute them by encoding the span using a character-based LSTM \cite{Hochreiter:1997:LSM:1246443.1246450}. The span vector $\vec{s}(m)$ is a concatenation of the word- and character-level vectors.   

\paragraph{Context.} The context surrounding a mention may indicate its compatibility with other candidate mentions  \cite{clark-manning:2016:P16-1,lee-EtAl:2017:EMNLP2017,kenyondean-cheung-precup:2018:S18-2}. To model context, we use ELMo, contextual representations derived from a neural language model \cite{peters-EtAl:2018:N18-1}. ELMo has recently improved performance on several challenging NLP tasks, including within-document entity coreference resolution \cite{lee-he-zettlemoyer:2018:N18-2}. We set the context vector $\vec{c}(m)$ to the contextual representation of $m$'s head word, taking the average of the 3 ELMo layers. 

\paragraph{Semantic dependency to other mentions.} \label{joint_features_1} To model dependencies between event and entity clusters, we identify semantic role relationships between their mentions using a semantic role labeling (SRL) system. 

For a given event mention $m_{v_i}$, we extract its arguments, focusing on 4 semantic roles of interest: \texttt{Arg0}, \texttt{Arg1}, \texttt{location}, and \texttt{time}. Consider a specific argument slot, e.g. \texttt{Arg1}. If the slot is filled with an entity mention $m_{e_j}$ which in the current configuration is assigned to an entity cluster $c$, we set the corresponding \texttt{Arg1} vector to the averaged span vector of all the mentions in $c$:  $\vec{d}_\text{\texttt{Arg1}}(m_{v_i}) = \frac{1}{|c|} \sum_{m \in c} \vec{s}(m)$. Otherwise we set $\vec{d}_\text{\texttt{Arg1}}(m_{v_i}) = \vec{0}$. The final vector $\vec{d}(m)$ is the concatenation of the various argument vectors: 
\vspace*{-8pt}
\begin{equation*}
\resizebox{0.91\hsize}{!}{%
    $\vec{d}(m_{v_i}) = [\vec{d}_\text{\texttt{Arg0}}(m_{v_i}); \vec{d}_\text{\texttt{Arg1}}(m_{v_i}); \vec{d}_\text{\texttt{loc}}(m_{v_i}); \vec{d}_\text{\texttt{time}}(m_{v_i})]$
    }
\vspace*{-5pt}
\end{equation*}

Symmetrically, we compute the argument vectors of an entity mention according to the events in which the entity mention plays a role.

This representation allows our model to directly compute the similarity between two mentions while considering a rich distributed representation of the current coreference clusters of their related arguments or predicates. \newcite{lee-EtAl:2012:EMNLP-CoNLL}, on the other hand,  modeled the dependencies between event and entity clusters using only simple discrete features, indicating the number of coreferring arguments across clusters.

\begin{figure}[!t]
\includegraphics[width=\columnwidth]{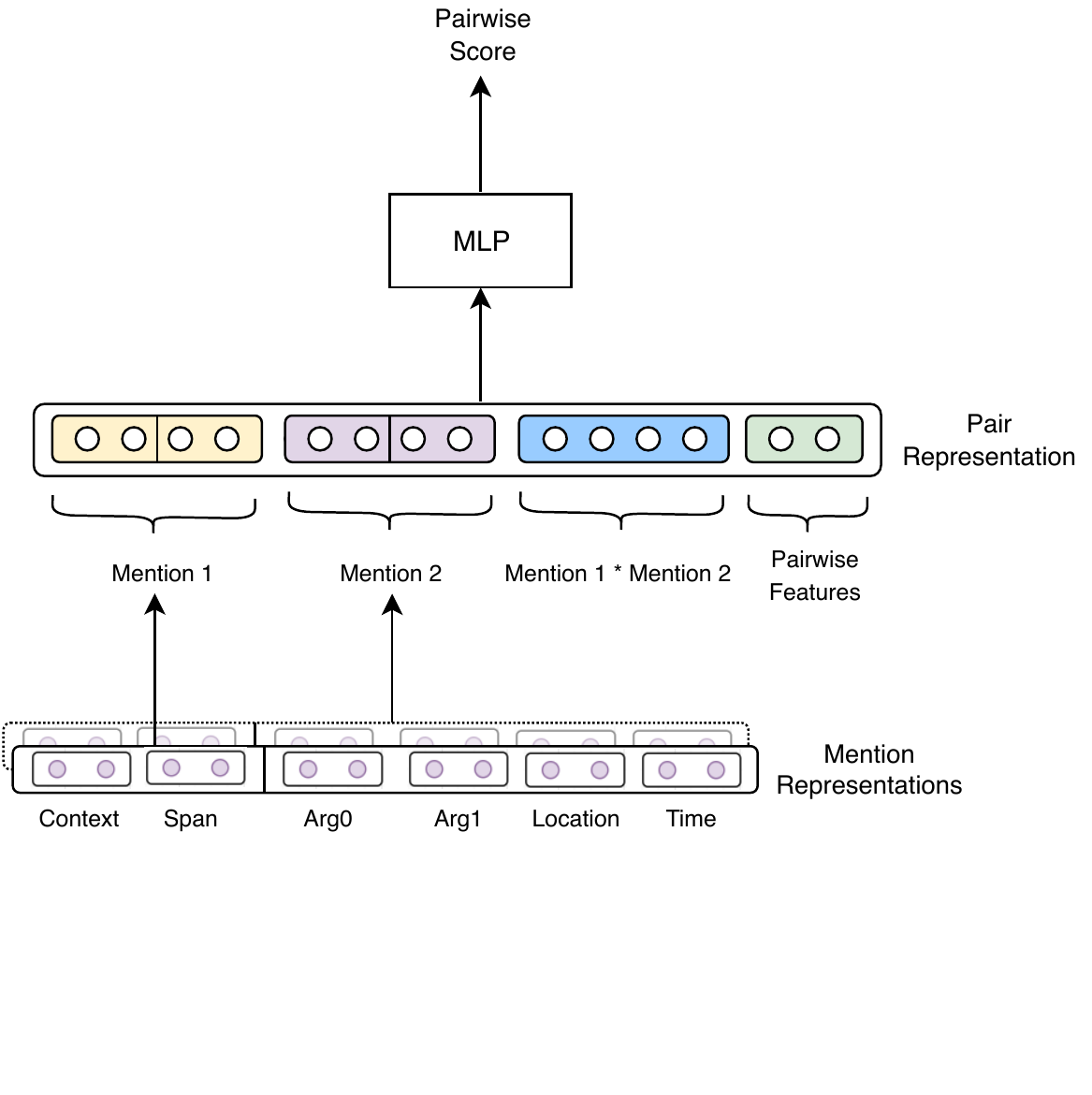}
\vspace{-70pt}
\caption{An illustration of the pairwise mention scorer. The bottom vectors are mention representations which include lexical and contextual features, and features derived from the mention's dependency on other mentions. The input to the network is a concatenation of two mention vectors with their element-wise multiplication and additional pairwise features.}
\label{fig:network}
\vspace{-7pt}
\end{figure}

The final mention vector is a concatenation of the various features: $\vec{v}(m) = [\vec{c}(m) ; \vec{s}(m) ; \vec{d}(m)]$, as illustrated in Figure~\ref{fig:network} (bottom row).

\subsection{Mention-Pair Coreference Scorer}
\label{sec:scoring}

Figure~\ref{fig:network} illustrates our pairwise mention scoring function $S(m_i, m_j)$ that returns a score denoting the likelihood that two mentions $m_i$ and $m_j$ are coreferring. We learn a separate function for entities ($S_E$) and for events ($S_V$), both trained identically as feed-forward neural networks. For the sake of simplicity, we describe them here as a single function $S(\cdot, \cdot)$.

The input to $S(m_i, m_j)$ is $\vec{v}_{i,j} =  [\vec{v}(m_i) ; \vec{v}(m_j) ; \vec{v}(m_i) \circ \vec{v}(m_j) ;  f(i,j)]$, where $\circ$ denotes an element-wise multiplication. Following \newcite{lee-EtAl:2012:EMNLP-CoNLL}, we enrich our mention-pair representation with four pairwise binary features $f(i,j)$, indicating whether the two mentions have coreferring arguments (or predicates) in a given role (\texttt{Arg0}, \texttt{Arg1}, \texttt{location}, and \texttt{time}). We encode each binary feature as 50-dimensional embedding to increase its signal. 

To train $S_E$ we take as training examples all pairs of entity mentions that belong to different entity clusters in the current predicted configuration $E_t$. The gold label for a given pair ($m_i$, $m_j$) is set to 1 if they belong to the same gold cluster, and to 0 otherwise. We train it using binary cross entropy as the loss function. $S_V$ is trained symmetrically.

\begin{figure*}[!t]
\input{figures/algorithms.tex}
\caption{Overview of the training algorithm (left) and the inference algorithm (right). The differences between the two procedures are highlighted.}
\label{alg:algorithm}
\end{figure*}       

\subsection{Inference}
\label{sec:inference}

Figure~\ref{alg:algorithm} describes our model step-by-step: the left part is the training procedure, while the right part is the inference procedure. The differences between the two procedures are highlighted. We first focus on the inference procedure (right), which gets as input the document set $D$, the pairwise mention scorers $S_E$ and $S_V$, and the gold standard mentions.\footnote{We follow the setup of \newcite{kenyondean-cheung-precup:2018:S18-2} and use the gold standard mentions (see Section~\ref{sec:experiments}).} 

The algorithm operates over each topic separately. To that end, we start by applying document clustering using the K-Means algorithm, yielding a set of topics T. For a given topic $t$, the algorithm uses the gold entity and event mentions to build initial clusters. Event clusters $V_t$ are initialized to singletons (line 2). Similarly to \newcite{lee-EtAl:2012:EMNLP-CoNLL}, entity clusters $E_t$ are initialized to the output of a within-document entity coreference resolution system (line 3).\footnote{This reduces the search space, and decouples cross-document entity resolution from the within-document variant. The latter consists of phenomena such as pronoun resolution that are already handled well by existing tools.} 
Our iterative algorithm alternates between entity and event clustering, incrementally constructing the final clustering configuration (lines 4-12). 

When the algorithm focuses on entities, it starts with updating the entity representations according to the event clusters in the current configuration, $V_t$ (line 6). This update includes the recreation of argument vectors for each entity mention, as described in Section~\ref{sec:representation}. We use agglomerative clustering that greedily merges multiple cluster pairs with the highest cluster-pair scores (line 8) until the scores are below a pre-defined threshold $\delta_2$. The algorithm starts with high-precision merges, leaving less precise decisions to a latter stage, when more information becomes available. We define the cluster-pair score as the average mention linkage score: $S_{cp}(c_i, c_j) = \frac{1}{|c_i| \cdot |c_j|} \cdot \sum_{m_i \in c_i}{\sum_{m_j \in c_j}{S(m_i, m_j)}}$. The same steps are repeated for events (lines 10-12), and repeat iteratively until no merges are available or up to a pre-defined number of iterations (line 4).  

\subsection{Training}
\label{sec:training}
The training steps are similarly described in the left part of Figure~\ref{alg:algorithm}. At each iteration, we train two updated scorer functions $S_E$ (line 7) and $S_V$ (line 11). Since our representation requires a clustering configuration, we use a training procedure that simulates the inference step. The training examples for each scorer change between iterations based on cluster-pair merges occurred in previous iterations. This allows our model to be trained on various predicted clustering configurations that are gradually improved during the training. 

The training procedure differs from the inference procedure by using the gold standard topic clusters and by initializing the entity clusters with the gold standard within-document coreference clusters. We do so in order to reduce the noise during training.

\subsection{Implementation Details} 
\label{sec:implementation_details}

Our model is implemented in PyTorch \cite{paszke2017automatic}, using the ADAM optimizer \cite{kingma2014adam} with a minibatch size of 16.
We initialize the word-level representations to the pre-trained 300 dimensional GloVe word embeddings \cite{pennington2014glove}, and keep them fixed during training. The character representations are learned using an LSTM with hidden size 50. We initialized them with pre-trained character embeddings\footnote{Available at \url{https://github.com/minimaxir/char-embeddings}}. Each scorer consists of a sigmoid output layer and two hidden layers with 4261 neurons activated by ReLU function \cite{nair2010rectified}. We set the merging threshold in the training step to $\delta_{1}=0.5$. We tune the threshold for inference step on the validation set to $\delta_{2}=0.5$.
To cluster documents into topics at inference time, we use the K-Means algorithm implemented in Scikit-Learn \cite{scikit-learn}. Documents are represented using TF-IDF scores of unigrams, bigrams, and trigrams, excluding stop words. We set $K=20$ based on the Silhouette Coefficient method \cite{rousseeuw1987silhouettes}, which successfully reconstructs the number of test sub-topics.
During inference, we use Stanford CoreNLP \cite{manning-EtAl:2014:P14-5} to initialize within-document entity coreference clusters.

\paragraph{Identifying Predicate-Argument Structures.} To extract relations between events and entities we follow previous work \cite{lee-EtAl:2012:EMNLP-CoNLL,TACL634,choubey-huang:2017:EMNLP20172} and extract predicate-argument structures using SwiRL \cite{surdeanu2007combination}, a semantic role labeling (SRL) system. To increase the coverage we apply additional heuristics:

\begin{itemize}[leftmargin=*,topsep=2pt,parsep=0pt,partopsep=0pt,itemsep=2pt]
    \item Since SwiRL only identifies verbal predicates, we follow \newcite{lee-EtAl:2012:EMNLP-CoNLL} and consider nominal event mentions with possesors (``\textit{Amazon's acquisition}'') as predicates and their \texttt{Arg0}.
    
    \item We use the spaCy dependency parser \cite{spacy2} to identify verbal event mentions whose subject and object are entities, and add those entities as their \texttt{Arg0} and \texttt{Arg1} roles, respectively. 

    \item Following \newcite{lee-EtAl:2012:EMNLP-CoNLL}, for a given event mention, we consider its closest left (right) entity mention as its \texttt{Arg0} (\texttt{Arg1}) role. 
\end{itemize}

\begin{table}[t!]
\small
\centering
\begin{tabular}{l l l l l}
\toprule 
& \textbf{Train} & \textbf{Validation} & \textbf{Test} & \textbf{Total} \\ \hline
\# Topics & 25 & 8 & 10 & 43\\
\# Sub-topics  & 50 & 16 & 20 & 86 \\
\# Documents  & 574 & 196 & 206 & 976 \\
\# Sentences & 1037 & 346 & 457 & 1840 \\
\# Event mentions  & 3808 & 1245 & 1780 & 6833\\
\# Entity mentions  & 4758 & 1476 & 2055 & 8289\\
\# Event chains & 1527 & 409 & 805 & 2741 \\
\# Entity chains  & 1286 & 330 & 608 & 2224\\
\bottomrule
\end{tabular}
\vspace{-8pt}
\caption{ECB+ statistics (including singleton clusters). The split to topics is as follows - Train: 1, 3, 4, 6-11, 13-17, 19-20, 22, 24-33; validation: 2, 5, 12, 18, 21, 23, 34, 35; test: 36-45.}
\label{data-stats-table}
\vspace{-10pt}
\end{table}

\section{Experimental Setup}
\label{sec:experiments}
\begin{table*}[t]
    \small
    \centering
    \begin{tabular}{l|ccc|ccc|ccc|c}
        \toprule
         & & MUC & & & B$^3$ & & & CEAF-$e$ & &  CoNLL \\
        \textbf{Model} & R & P & $F_1$ & R & P & $F_1$ & R & P & $F_1$ & $F_1$\\
        \hline
        \textsc{Cluster+Lemma} &71.3&83&76.7&53.4&84.9&65.6&70.1&52.5&60&67.4 \\
        \textsc{Disjoint} &76.7&80.8&78.7&63.2&78.2& 69.9&65.3&58.3&61.6&70\\
        \textsc{Joint~~~~~~~~~~~~~~~~~~~~~~~~~~~~~~~~~~~~~~~~~~~~~~} &78.6&80.9&79.7&65.5&76.4&70.5&65.4&61.3&63.3&\textbf{71.2}\\
        \bottomrule
        
    \end{tabular}
    \vspace*{-7pt}
    \caption{Combined within- and cross-document entity coreference results on the ECB+ test set.}
    \label{entity_results}
    \vspace*{-7pt}
\end{table*}

\begin{table*}[t]
    \small
    \centering
    \begin{tabular}{l|ccc|ccc|ccc|c}
        \toprule
         & & MUC & & & B$^3$ & & & CEAF-$e$ & &  CoNLL \\
        \textbf{Model} & R & P & $F_1$ & R & P & $F_1$ & R & P & $F_1$ & $F_1$\\
        \hline
        \textbf{Baselines} &&&&&&&&&&\\
        \textsc{Cluster+Lemma} &76.5&79.9&78.1&71.7&85&77.8&75.5&71.7&73.6& 76.5\\
        \textsc{CV} \cite{cybulska2015bag} &71&75&73&71&78&74&-&-&64& 73 \\
        \textsc{KCP} \cite{kenyondean-cheung-precup:2018:S18-2} &67&71&69&71&67&69&71&67&69&69\\
        \textsc{Cluster+KCP}  &68.4&79.3&73.4&67.2&87.2&75.9&77.4&66.4&71.5&73.6\\
        \hline
        
        \textbf{Model Variants} &&&&&&&&&&\\
        \textsc{Disjoint} &75.5&83.6&79.4&75.4&86&80.4&80.3&71.9&75.9&78.5\\
        \textsc{Joint} &77.6&84.5&80.9&76.1&85.1&80.3&81&73.8&77.3&\textbf{79.5}\\
        \bottomrule
        
    \end{tabular}
    \vspace*{-7pt}
    \caption{Combined within- and cross-document event coreference results on the ECB+ test set.}
    \label{event_results}
    \vspace*{-7pt}
\end{table*}
 
\begin{table}[t!]
\newcolumntype{Y}{>{\centering\arraybackslash}X}
\newcommand{\colindent}{\;}
\setlength{\tabcolsep}{0.25em}
\centering
\small
\begin{tabularx}{\linewidth}{l Y Y}
\toprule
 & CoNLL $F_1$ & $\Delta$ \\
\midrule
Joint model & 79.5 & \\
\colindent $-$ Pairwise binary features  & 79.4 & -0.1 \\
\colindent $-$ Dependent mentions vectors  & 78.6 & -0.9 \\
\colindent $-$ Both & 78.5 & -1.0 \\
\bottomrule
\end{tabularx}
\vspace{-7pt}
\caption{Ablations of the joint modeling parts in our architecture. CoNLL $F_1$ score is reported for combined within- and cross-document event coreference.}
\label{tab:ablations}
\vspace{-7pt}
\end{table}

We use the ECB+ corpus, which is the largest dataset consisting of within- and cross-document coreference annotations for entities and events. 

We follow the setup of \newcite{cybulska-vossen:2015:EVENTS}, which was also employed by \newcite{kenyondean-cheung-precup:2018:S18-2}. This setup uses a subset of the annotations which has been validated for correctness by \newcite{cybulska2014using} and allocates a larger portion of the dataset for training (see Table~\ref{data-stats-table}). Since the ECB+ corpus only annotates a part of the mentions, the setup uses the gold-standard event and entity mentions rather, and does not require specific treatment for unannotated mentions during evaluation.

A different setup was carried out by \newcite{TACL634} and \newcite{choubey-huang:2017:EMNLP20172}. They used the full ECB+ corpus, including parts with known annotation errors. At test time, they rely on the output of a mention extraction tool  \cite{TACL634}. To address the partial annotation of the corpus, they only evaluated their systems on the subset of predicted mentions which were also gold mentions. Finally, their evaluation setup was criticized by \newcite{upadhyay-EtAl:2016:COLING} for ignoring singletons (cluster with a single mention), effectively making the task simpler; and for evaluating each sub-topic separately, which entails ignoring incorrect coreference links across sub-topics. 

\paragraph{Evaluation Metrics.} We use the official CoNLL scorer \cite{pradhan-EtAl:2014:P14-2},\footnote{\href{http://conll.github.io/reference-coreference-scorers/}{http://conll.github.io/reference-coreference-scorers/}} and report the performance on the common coreference resolution metrics: \textbf{MUC} \cite{vilain1995model},  $\mathbf{B^3}$ \cite{bagga1998algorithms}, \textbf{CEAF-}$\mathbf{e}$ \cite{luo:2005:HLTEMNLP}, and \textbf{CoNLL} $\mathbf{F_1}$, the average of the 3 metrics.

\section{Baselines}
\label{sec:baselines}

We compare our full model to published results on ECB+, available for event coreference only, as well as to a disjoint variant of our model and a deterministic lemma baseline.\footnote{We do not compare our work to \newcite{TACL634} and \newcite{choubey-huang:2017:EMNLP20172}, since they used another incomparable evaluation setup, as discussed in Section \ref{sec:experiments}.} 

\paragraph{\textsc{Cluster+Lemma}.} We first cluster the documents to topics (Section~\ref{sec:inference}), and then group mentions within the same document cluster which share the same head lemma.
This baseline differs from the lemma baseline of \newcite{kenyondean-cheung-precup:2018:S18-2} which is applied across topics.

\paragraph{\textsc{CV}} \cite{cybulska2015bag} is a supervised method for event coreference, based on discrete features. They first cluster documents to topics, and then cluster coreferring mentions within each topic cluster. Events are represented using information about participants, time and location, while documents are represented as ``bag-of-events''. We compare to their best reported results, differing from the CV baseline in \newcite{kenyondean-cheung-precup:2018:S18-2} which refers to the partial model that uses the same annotations in terms of sub-components of the event structure.

\paragraph{\textsc{KCP}} \cite{kenyondean-cheung-precup:2018:S18-2} is a neural network-based model for event coreference. They encode an event mention and its context into a vector and use it to cluster mentions. The model does not cluster documents to topics as a pre-processing step, but instead encodes the document as part of the mention representation, aiming to avoid spurious cross-topic coreference links thanks to distant document representations.

\paragraph{\textsc{Cluster+KCP}} To tease apart the contribution of our document clustering component from that of the rest of the model, we add a variant of the \textsc{KCP} model which relies on our document clustering component as a pre-processing step. During inference, we restrict their model to clustering mentions only within the same document cluster. Accordingly, we re-trained their model using the gold document clusters for hyper-parameters tuning to fit this cluster-based setting.

\paragraph{\textsc{Disjoint}.} A variant of our model which uses only the span and context vectors to build mention pair representations, ablating joint features. 

\paragraph{}We do not compare our work directly to \newcite{lee-EtAl:2012:EMNLP-CoNLL} since it was evaluated on a different corpus and using a different evaluation setup. Instead, we compare to CV and KCP, more recent models which reported their results on the ECB+ dataset.

With respect to entity coreference, to the best of our knowledge, our work is the first to publish entity coreference results on the ECB+ dataset. We therefore only compare our performance to that of the lemma baseline and our disjoint model.  
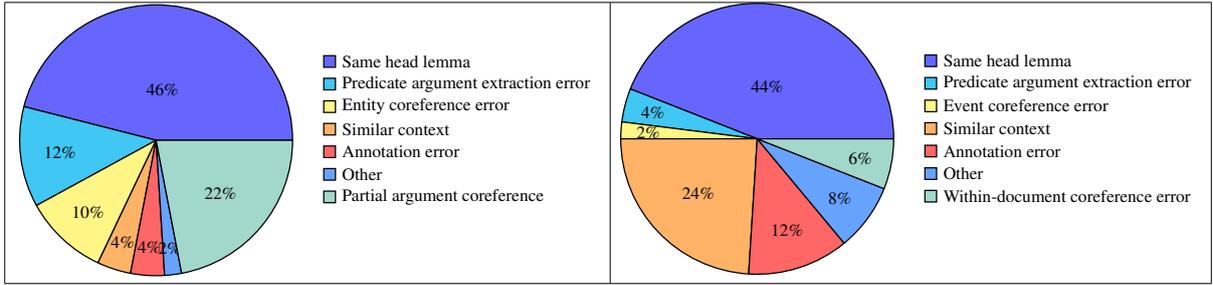
\begin{figure*}[!t]
    \centering
    \resizebox{1\textwidth}{!}{
    \begin{tabular}{|l|l|}
    \hline
    \begin{tikzpicture}
\pie[text=legend]{46/Same head lemma, 12/Predicate argument extraction error, 10/Entity coreference error, 4/Similar context, 4/Annotation error, 2/Other, 22/Partial argument coreference}
\end{tikzpicture} & 
    \begin{tikzpicture}
\pie[text=legend]{44/Same head lemma, 4/Predicate argument extraction error, 2/Event coreference error, 24/Similar context, 12/Annotation error, 8/Other, 6/Within-document coreference error}
\end{tikzpicture} \\ \hline
    \end{tabular}
    }
    \vspace*{-7pt}
    \caption{Event coreference errors (left) and entity coreference errors (right).}
    \label{fig:error_analysis}
    \vspace*{-7pt}
\end{figure*}

\section{Results}
\label{sec:results}
Table~\ref{entity_results} presents the performance of our method with respect to entity coreference. Our joint model improves upon the strong lemma baseline by 3.8 points in CoNLL $F_1$ score.

Table~\ref{event_results} presents the results on event coreference. Our joint model outperforms all the baselines with a gap of 10.5 CoNLL $F_1$ points from the last published results (\textsc{KCP}), while surpassing our strong lemma baseline by 3 points. 

The results reconfirm that the lemma baseline, when combined with effective topic clustering, is a strong baseline for CD event coreference resolution on the ECB+ corpus \cite{upadhyay-EtAl:2016:COLING}. In fact, thanks to our near-perfect topic clustering on the ECB+ test set (Homogeneity: 0.985, Completeness: 0.982, V-measure: 0.984, Adjusted Rand-Index: 0.965), the \textsc{Cluster+Lemma} baseline surpasses prior results on ECB+. 

The results of \textsc{Cluster+KCP} again indicate that pre-clustering of documents to topics is beneficial, improving upon the \textsc{KCP} performance by 4.6 points, though still performing substantially worse than our joint model. 

To test the contribution of joint modeling, we compare our joint model to its disjoint variant. We observe that the joint model performs better on both event and entity coreference. The performance gap is modest but significant with bootstrapping and permutation tests ($p < 0.001$).    

We further ablate additional components from the full representation (Table \ref{tab:ablations}). We show that each of our representation components contributes to performance, but the continuous vector components representing semantic dependency to other mentions are stronger than the pairwise binary features originally used by \newcite{lee-EtAl:2012:EMNLP-CoNLL}.

\section{Analysis}
\label{sec:analysis}
\begin{figure}[!t]
\vspace{5pt}
    \centering
    \resizebox{0.4\textwidth}{!}{
    \begin{tabular}{l}
    \includegraphics[trim={36pt 25pt 30pt 10pt},clip]{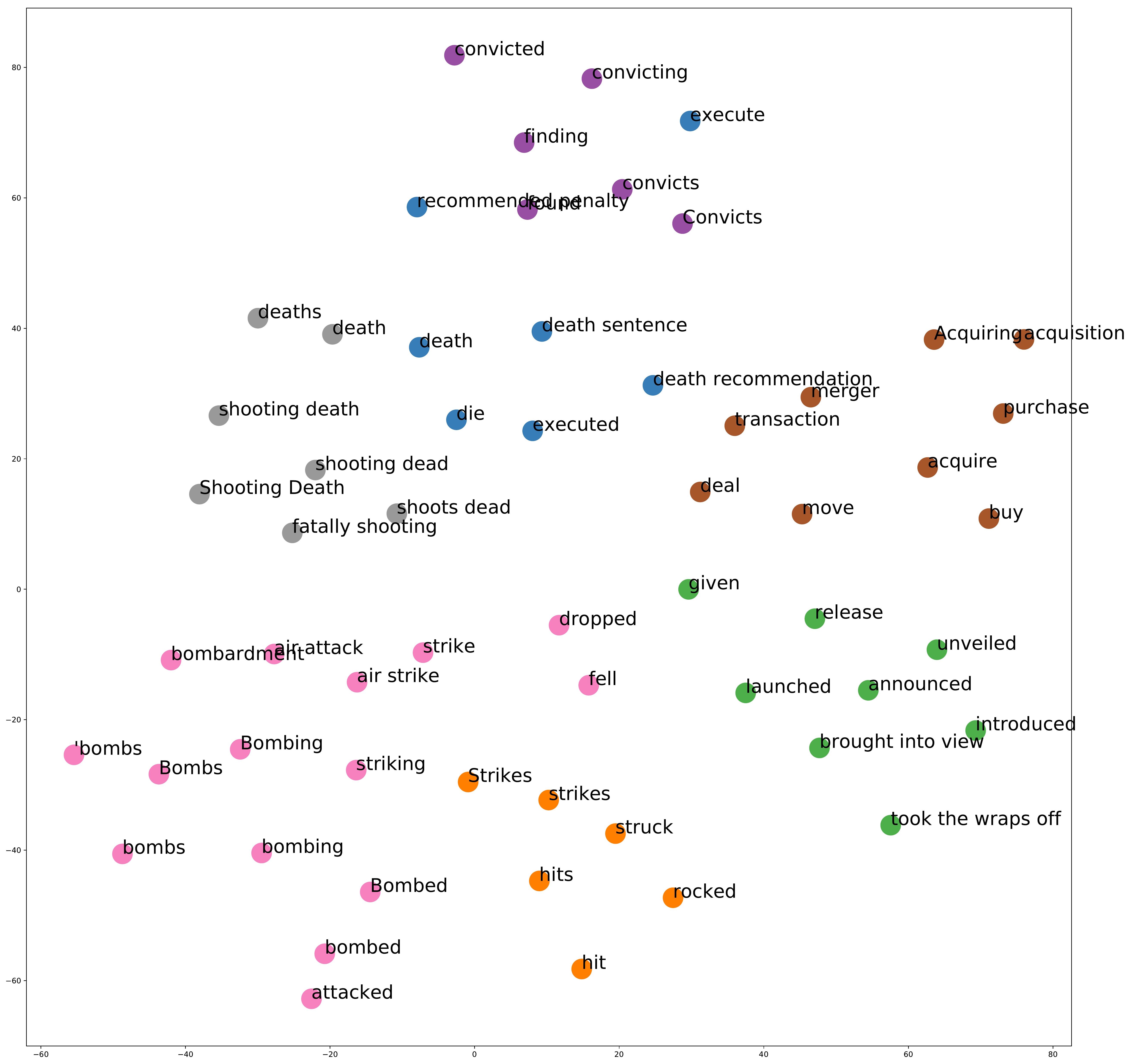} \\ 
    \includegraphics[trim={36pt 30pt 30pt 10pt},clip]{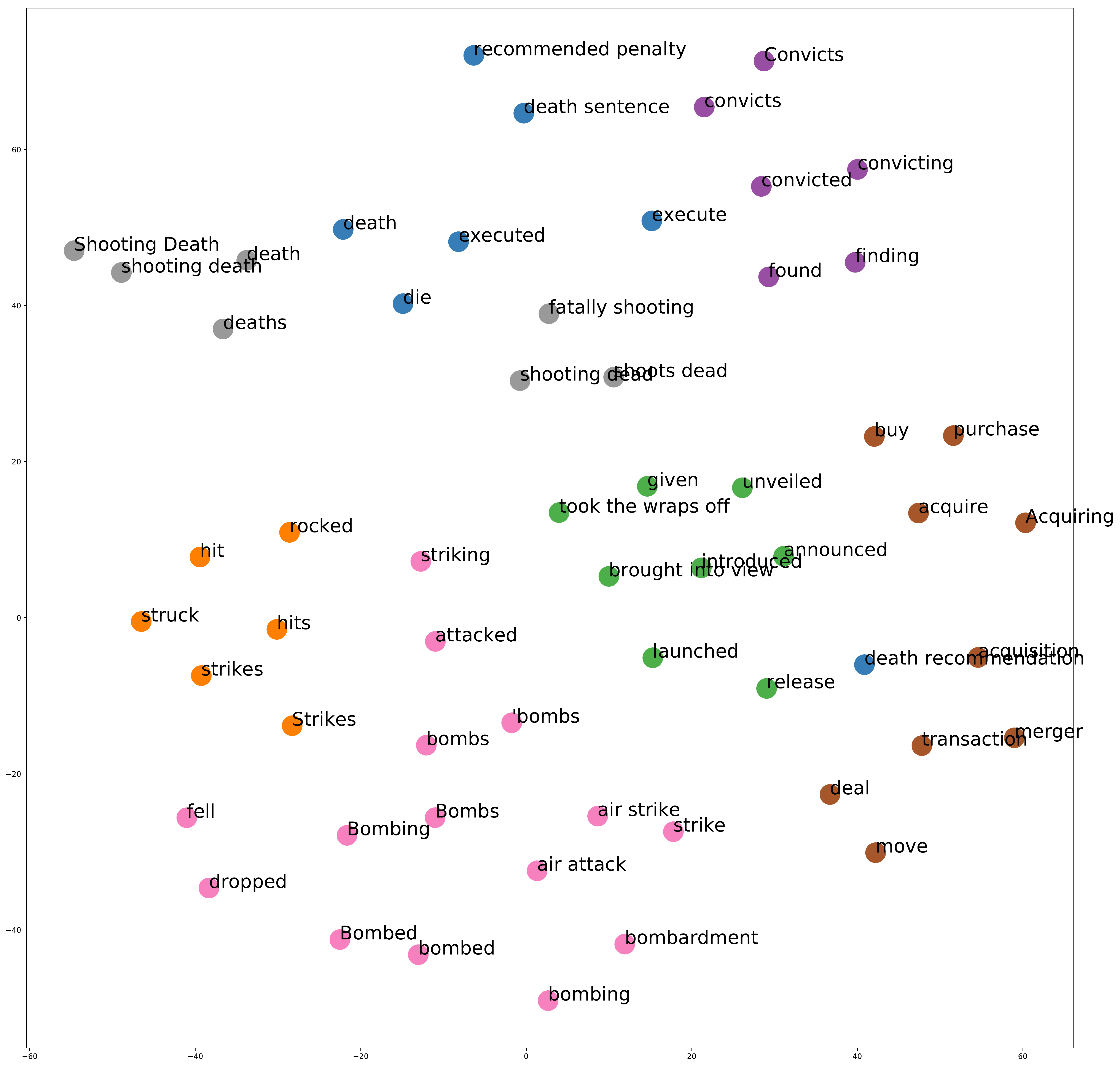} \\ 
    \includegraphics[trim={36pt 10pt 10pt 10pt},clip]{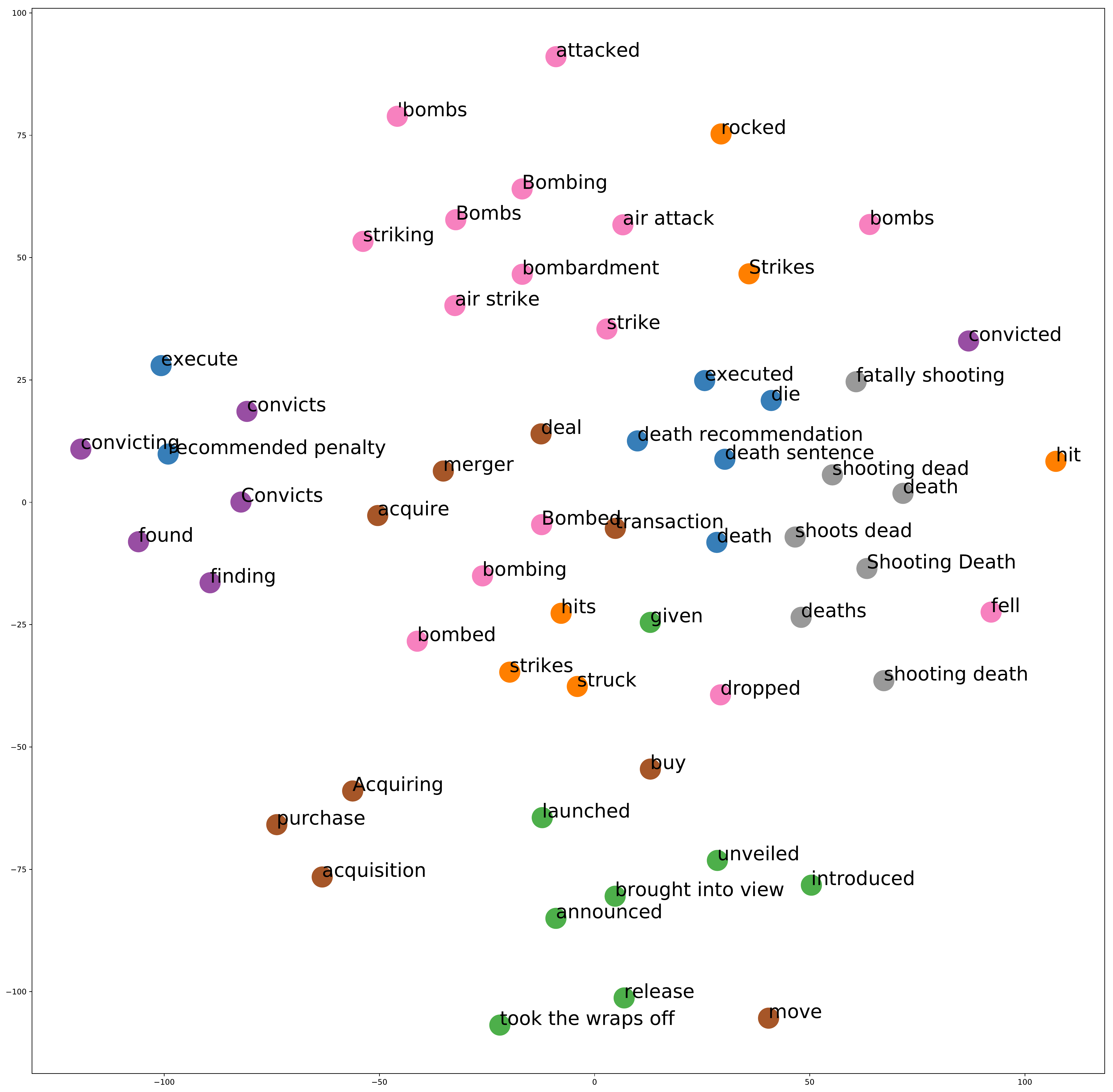} \\ 
    \end{tabular}
    }
    \vspace{-10pt}
    \caption{t-SNE projection of the full mention representation (top), context vector (middle) and dependent mention vector (bottom). Each point is an event mention, colored according to its gold cluster.}
    \label{fig:tsne_analysis}
    \vspace{-12pt}
\end{figure} 

\subsection{Error Analysis}
\label{sec:error_analysis}

To analyze the errors made by our joint model we sampled 50 event mentions and 50 entity mentions that were clustered incorrectly, i.e. where their predicted cluster contained at least 70\% of mentions that are not in their gold cluster. 

Figure~\ref{fig:error_analysis} shows a pie chart for each mention type, manually categorized to error types, suggesting future areas for improvement. For both entities and events, mentions were often clustered incorrectly with other mentions that share the same head lemma. Errors in the extraction of the predicate-argument structures accounted for 12\% of the errors in events and 4\% for entities, e.g. marking \textit{dozens} as the \texttt{Arg0} of \textit{devastated} in  ``\textit{dozens in a region devastated by the quake}''. 

The joint features caused 10\% of the event errors and 2\% of the entity errors, where two non-coreferring event mentions were clustered to the same event cluster based on their entity arguments that were incorrectly predicted as coreferring, and vice versa. For example, the event \textit{shakes} in ``\textit{earthquake shakes Lake County}'' and ``\textit{earthquake shakes Northern California}'' was affected by the wrong coreference clustering of ``\textit{Lake County}'' and ``\textit{Northern California}''. 

We also found mentions that were wrongly clustered together based on contextual similarity (24\% for entities, 4\% for events) as well as some annotation errors (12\% and 4\%). The within-document entity coreference system caused additional 6\% of entity errors. Finally, 22\% of the event errors were caused by event mentions sharing coreferring arguments. This may happen for instance when similar events occur at different times (``\textit{The earthquake struck at about \textbf{9:30} a.m. and had a depth of \textbf{2.7} miles, according to the \textbf{USGS}.}'' vs. ``\textit{The earthquake struck at about \textbf{7:30} a.m. and had a depth of \textbf{1.4} miles, according to the \textbf{USGS}.}''). 

\subsection{Mention Representation Components}
\label{sec:mention_rep_components}
To understand the contribution of each component in the mention representation to the clustering, we visualize them. We focus on events, and sample 7 gold clusters from the test set that have at least 5 mentions each. We then compute t-SNE projections \cite{maaten2008visualizing} of the full mention representation, only the context vector, and only the semantically-dependent mentions vector (top, middle, and bottom parts of Figure~\ref{fig:tsne_analysis}). 
In all the 3 graphs, each point refers to an event mention and its color represents the mention's gold cluster. The full mention representations (top) yield visibly better clusters, but the context vectors (middle) are also quite accurate, emphasizing the importance of modeling context for resolving coreference. The semantically-dependent mentions vectors (bottom) are less accurate on their own, yet, they manage to separate well some clusters even without access to the mention span itself, and based only on the predicate-argument structures.

\section{Conclusion}
\label{sec:conclusion}
We presented a neural approach for resolving cross-document event and entity coreference. We represent a mention using its text, context, and---inspired by the joint model of \newcite{lee-EtAl:2012:EMNLP-CoNLL}---we make an event mention representation aware of coreference clusters of entity mentions to which it is related via predicate-argument structures, and vice versa. Our model achieves state-of-the-art results, outperforming previous models by 10.5 CoNLL $F_1$ points on events, and providing the first cross-document entity coreference results on ECB+. Future directions include investigating ways to minimize the pipeline errors from the extraction of predicate-argument structures, and incorporating a mention prediction component, rather than relying on gold mentions.

\newpage

\section*{Acknowledgments}

We would like to thank Jackie Chi Kit Cheung for the insightful comments. This work was supported in part by an Intel ICRI-CI grant, the Israel Science Foundation grant 1951/17, the German Research Foundation through the German-Israeli Project Cooperation (DIP, grant DA 1600/1-1), and a grant from Reverso and Theo Hoffenberg.

\bibliography{references}
\bibliographystyle{acl_natbib}

\end{document}